\documentclass[final]{cvpr}

\usepackage{times}
\usepackage{epsfig}
\usepackage{graphicx}
\usepackage{amsmath}
\usepackage{amssymb}
\usepackage{color}
\usepackage{amsmath}
\usepackage{arydshln}
\usepackage[switch]{lineno}
\graphicspath{{images/}}

\usepackage[marginal]{footmisc}

\usepackage[pagebackref=true,breaklinks=true,colorlinks,bookmarks=false]{hyperref}
\newcommand{\Tref}[1]{Table~\ref{#1}}
\newcommand{\Eref}[1]{Equation~(\ref{#1})}
\newcommand{\Fref}[1]{Figure~\ref{#1}}

\def\confYear{CVPR 2021}

\begin{document}

\title{Improved Image Matting via Real-time User Clicks and Uncertainty Estimation}

\author{\Large Tianyi Wei\textsuperscript{\rm 1}, Dongdong Chen\textsuperscript{\rm 2,}\textsuperscript{$\dagger$}, Wenbo Zhou\textsuperscript{\rm 1,}\textsuperscript{$\dagger$}, Jing Liao\textsuperscript{\rm 3},\\
	 \Large Hanqing Zhao\textsuperscript{\rm 1},  Weiming Zhang\textsuperscript{\rm 1}, Nenghai Yu\textsuperscript{\rm 1} \\
 \textsuperscript{\rm 1}University of Science and Technology of China  \ \textsuperscript{\rm 2}Microsoft Cloud AI  \   \\ 
  \textsuperscript{\rm 3}City University of Hong Kong\\
  {\tt\small\{bestwty@mail., welbeckz@, zhq2015@mail., zhangwm@, ynh@ \}.ustc.edu.cn } \\
{\tt\small cddlyf@gmail.com}, {\tt\small jingliao@cityu.edu.hk}
}

\maketitle

\footnote{\textsuperscript{$\dagger$}\  Corresponding Author.}
\footnote{The demo of our real-time user interactive matting system can be find at \url{https://youtu.be/pAXydeN-LpQ}.}

\begin{abstract}
   Image matting is a fundamental and challenging problem in computer vision and graphics. Most existing matting methods leverage a user-supplied trimap as an auxiliary input to produce good alpha matte. However, obtaining high-quality trimap itself is arduous, thus restricting the application of these methods. Recently, some trimap-free methods have emerged, however, the matting quality is still far behind the trimap-based methods. The main reason is that, without the trimap guidance in some cases, the target network is ambiguous about which is the foreground target. In fact, choosing the foreground is a subjective procedure and depends on the user's intention. To this end, this paper proposes an improved deep image matting framework which is trimap-free and only needs several user click interactions to eliminate the ambiguity. Moreover, we introduce a new uncertainty estimation module that can predict which parts need polishing and a following local refinement module. Based on the computation budget, users can choose how many local parts to improve with the uncertainty guidance. Quantitative and qualitative results show that our method performs better than existing trimap-free methods and comparably to state-of-the-art trimap-based methods with minimal user effort.
\end{abstract}

\section{Introduction}

Image matting refers to the task of precisely separating the foreground object from the background and accurately estimating the per-pixel opacity near the boundary. It has been studied by academic and industrial communities for many years. Typical applications include image editing, film production and virtual background for video conferencing. Given an input image $ I $, it can be formulated as a mathematical optimization problem as follows:
\begin{equation}
	I_{i} = {\alpha}_{i}{F}_{i} + (1-{\alpha}_{i}){B}_{i},\label{mattingequation}
\end{equation}
where $ \alpha_{i}\in[0,1] $ denotes the opacity of the foreground object at pixel $ i $. It can be observed that, for each pixel, this problem needs to solve 7 unknown values from only 3 known values. Therefore, it is a highly ill-posed problem.

To address this problem, many classical algorithms \cite{Aksoy2017DesigningEI,Lee2011NonlocalM,Chuang2001ABA,Wang2007OptimizedCS,Xu2017DeepIM,Chen2018TOMNetLT} have been proposed by leveraging a well-defined trimap to constrain the solution space. The trimap divides the image into three types of areas: foreground, background, and transition regions. 
Then the matting task is then simplified as the problem of estimating the unknown values only in the transition region. Based on this simplification, such trimap-based methods \cite{Li2020NaturalIM,Hou2019ContextAwareIM,Cai2019DisentangledIM} can usually achieve very good performance. However, drawing a suitable trimap itself is very tedious and time-consuming. For some complex cases, it will even cost more than 10 minutes. Therefore, it is not that friendly, especially to non-professional users. 

With the development of deep learning, some studies\cite{Chen2018SemanticHM,Zhang2019ALF,Sengupta2020BackgroundMT} that do not require the trimap input emerge recently. However, their performance is still far behind the trimap-based methods. The main reason is that, without the trimap guidance in some cases, deep networks become ambiguous about which is the foreground target to process. To alleviate the ambiguity, a large-scale matting dataset of a certain target category (e.g., portrait) is often required for training. But this solution is not scalable and expensive. More importantly, it will totally fail if the users want to choose some new categories that do not appear in the training set. In our understanding, choosing the foreground is subjective and thus some user hint is unavoidable. And the key challenge is how to minimize the user effort in the interactions.

To overcome the aforementioned limitations, we propose a novel matting approach that considers both performance and user-friendliness, by introducing user clicks into the matting network. To make the network adaptive to any user clicks, we simulate user clicks during training by adding random points into the foreground and background respectively. Compared with trimap-based schemes, our method only requires a few user clicks as hints to indicate the foreground and background but can achieve comparable performance. And in most cases which has no foreground ambiguity problem, ours even does not need any user clicks. When compared to trimap-free methods, our method does not suffer from the ambiguity problem and performs much better. More importantly, it is easy to extend to any unseen category by adding only a few user clicks.

Besides, we further introduce a new uncertainty estimation module and a corresponding uncertainty-guided local refinement network. Parallel to the main alpha matte estimation branch, the uncertainty estimation branch can predict which parts need further refinement. Based on the computation budget, the users can choose how many local parts to improve with the uncertainty guidance. Compared to some existing methods that use an extra network for global refinement, this manner is more flexible and efficient by avoiding some redundant computation for the well-predicted regions. To the best of our knowledge, we are the first that introduces uncertainty in deep image matting.

To demonstrate the effectiveness of our method, we conduct extensive experiments on both synthetic and real datasets and show superior performance. Ablation analysis also justifies its flexibility and adaptability to new categories unseen in the training set. For better application, a real-time user interactive system is provided.

To summarize, our contributions are three-fold as below:
\begin{itemize}
	\item We present the first attempt that introduces user click interaction into the image matting task. It can effectively eliminate the ambiguity and boost the matting performance with minimal user effort. In this sense, it can be seen as a new matting scheme between trimap-based and trimap-free methods.
	
	\item  We introduce a novel uncertainty estimation module and a corresponding uncertainty-guided local refinement network. By using the uncertainty map as hints, the network can perform automatic local refinement to produce more precise details and remove undesired blurring artifacts. 
	
	\item  The experimental results show our method performs considerably better than trimap-free approaches and comparably to state-of-the-art trimap-based methods. Besides, a real-time user interactive system is built.
\end{itemize}

\section{Related Work}
\vspace{0.3em} \noindent\textbf{Natural Image Matting.}  Existing image matting algorithms can be broadly categorized into two types: traditional prior-based and learning-based. Besides an input image, both types of algorithms may take some auxiliary inputs, including scribbles \cite{wang2005iterative}, trimaps\cite{Chuang2001ABA} or accumulated trimaps\cite{Yang2018ActiveM} to improve matting quality. 

For traditional prior-based methods, they can be further divided into color sampling-based methods and alpha propagation-based methods. Based on the local smoothness assumption of image statistics, sampling-based methods \cite{Chuang2001ABA, Feng2016ACS, Gastal2010SharedSF, He2011AGS}  usually first model the foreground and background color distributions then solve \Eref{mattingequation} to get the target opacity. Compared to sampling-based methods, propagation-based methods \cite{Aksoy2017DesigningEI, Chen2013KNNM, Levin2008ACS, Sun2004PoissonM} can potentially avoid the matte discontinuities issue. In detail, by utilizing the affinities between neighboring pixels, they propose to propagate the alpha values from known regions into unknown ones.

In recent years, deep learning-based algorithms have shown superior performance in image matting tasks. With the synthetic image matting dataset Deep Image Matting\cite{Xu2017DeepIM}, many data-driven works have been proposed. For example, Cai et al.  \cite{Cai2019DisentangledIM} decouple the matting task into two sub-tasks: trimap adaptation and alpha estimation. Inspired by traditional sample-based methods, Tang et al.  \cite{Tang2019LearningBasedSF} propose to first estimate the foreground and background color before estimating the alpha matte.  Inspired by the affinity-based method and the successes of contextual attention in inpainting, GCA-matting\cite{Li2020NaturalIM} introduces a guided contextual attention module into the network and achieves state-of-the-art performance. However, all these methods are trimap-based and need laborious trimap acquisition.

More recently, some trimap-free methods \cite{Zhang2019ALF,Liu2020BoostingSH} emerge. LF-Matting \cite{Zhang2019ALF} proposes a two-stage matting framework and generates trimap implicitly through the first stage. And in BoostingNet\cite{Liu2020BoostingSH}, Liu et al.  aim to strengthen the matting performance specifically for humans through utilizing both coarse and fine annotations. But due to the lack of the trimap prior, these trimap-free methods suffer from the ambiguity problem or cannot extend to new categories unseen during training. Therefore, their performance is still worse than trimap-based methods. Inspired by the interactive colorization work\cite{Zhang2017RealtimeUI}, our method innovatively introduces the click-based interaction into the image matting task, which resolves both  ambiguity and generality issues with very minimal user effort. Besides, we are the first that introduces the uncertainty-guided local refinement mechanism, which is more flexible and efficient than global refinement used in previous methods.

\noindent\textbf{Uncertainty Estimation in DNNs.} Uncertainty estimation is important for evaluating the robustness of deep learning models in computer vision. It is especially useful in some security-sensitive application scenarios because it can tell the users how confident deep models make the final prediction. Generally, existing uncertainty estimation methods used in DNNs can be grouped into two classes: sampling-based and sampling-free approaches. Typical sample-based methods include Monte Carlo dropout\cite{Gal2016DropoutAA}, Bayesian neural networks\cite{Mandt2017StochasticGD,Shridhar2019ACG}, and Bootstrapped ensembles\cite{Lakshminarayanan2017SimpleAS}.  They often rely on multiple evaluations for the same input to obtain the uncertainty and bootstrapped ensembles also need to store several sets of weights. Compared to sampling-based methods, sampling-free methods \cite{Lakshminarayanan2017SimpleAS,Ilg2018UncertaintyEA,Meyer2019LaserNetAE} directly model the prediction target as a Gaussian or Laplacian distribution rather than a deterministic value, then optimize the distribution parameters with a bayesian-like formulation. In this paper, we adopt a sampling-free manner in the uncertainty estimation module due to its simplicity.

\section{Proposed Method}
\subsection{Motivation}
As mentioned before, both trimap-based and trimap-free methods have their advantages and disadvantages. Specifically, trimap-based methods can achieve state-of-the-art performance, but they need users to supply a well-drawn trimap to indicate the foreground, background and transition regions. Empirically, by using advanced commercial systems like Photoshop, even a professional user still needs several minutes to draw a suitable trimap. By contrast, trimap-free methods do not require any user prior but their performance is significantly worse. This is because, without the prior, such methods are often ambiguous about which are the target foreground objects, especially in some complex cases. To address the ambiguity issue, one typical solution would be collecting a large scale labeled dataset for one interested category so that the network can rely on the semantics to identify the foreground. However, this solution is not scalable enough because data labeling  is expensive and it cannot extend to unseen categories, like the \textit{``surf board"} in the first case of \Fref{fig:realimage}. Besides, even for one specific category, it still cannot satisfy users' needs in some cases.  For example, in the second case of \Fref{fig:realimage}, users may only want to keep one of the target portraits.

Based on the above analysis, we can conclude that: 1) The user prior is a very important hint to reduce the ambiguity for image matting and extend to unseen new categories. 2) Trimap is a very strong prior, but its acquisition is too time-consuming. This motivates us to refind one simpler user prior which is enough to identify the foreground object with minimal effort. To determine the prior type, let us recall some typical scenarios happening around us everyday. For example, imagine we are buying coffee in Starbucks, and we have a menu and want to order one specific coffee type. Apart from directly saying the coffee name, the most popular interaction way is simply using the finger to point the picture/name on the menu. Inspired by this natural interaction manner, we present the first attempt that uses the simple user clicks as the matting prior in this paper.

Besides, as demonstrated in many previous works \cite{Xu2017DeepIM,Zhang2019ALF}, an extra refinement stage can further improve the matting quality. They directly do it by feeding the first-stage matting result together with the original image  into the second refinement stage, which we call ``global refinement".  In this paper, we argue such a ``global refinement" scheme is neither effective nor flexible. Therefore, we further introduce a new uncertainty estimation module that can automatically predict which local parts need more polishing. With the uncertainty guidance and computation budget, users can flexibly choose which and how many parts to improve.

\begin{figure*}[t]
	\centering
	\includegraphics[width=\textwidth]{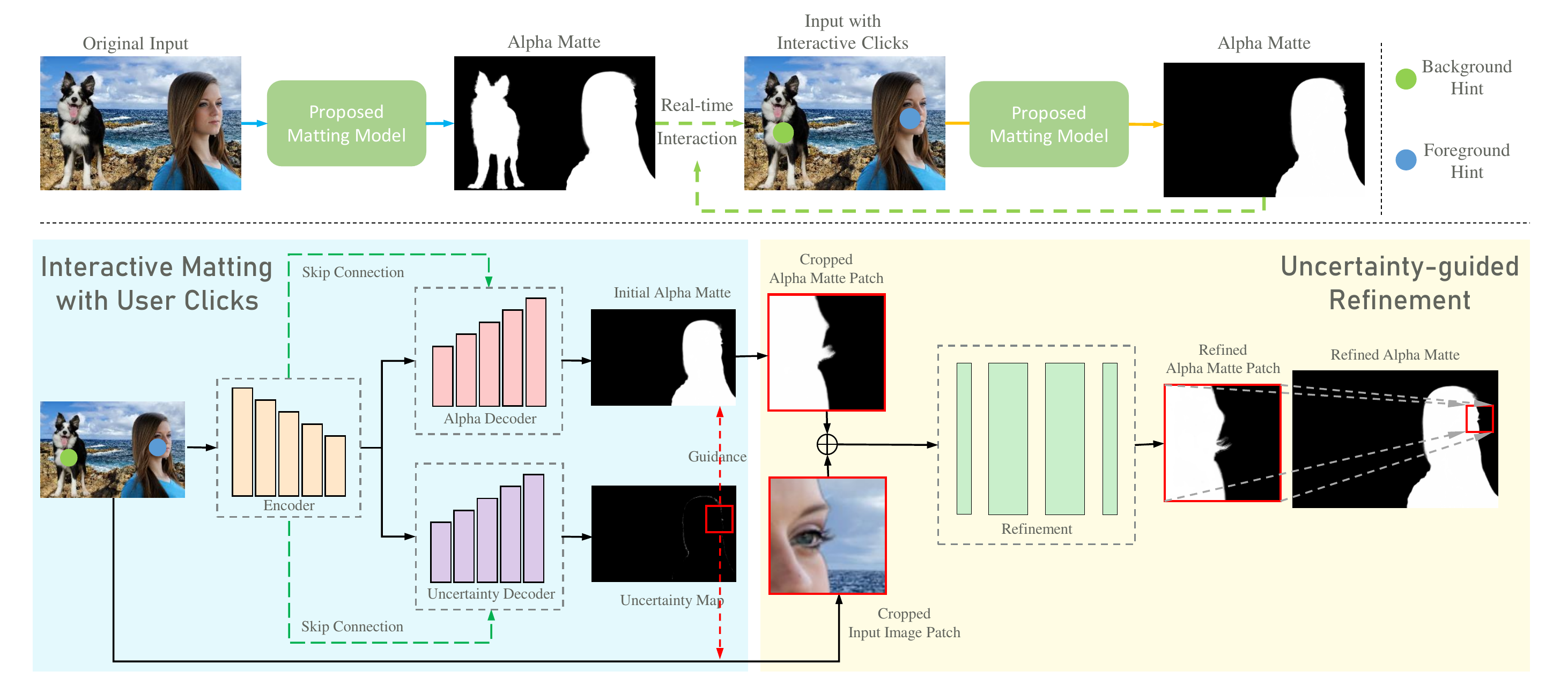} 
	\caption{Top: the overview of our newly proposed real-time interactive matting framework. Rather than relying on the strong trimap prior, only several user clicks are needed to specify the desired foreground and background. Bottom: the detailed architecture consists of two key components: the interactive matting network and the uncertainty-guided local refinement module.} 
	\label{fig:networkfig}
\end{figure*}

By combining these two motivations together, we propose a new matting framework as shown in \Fref{fig:networkfig}. It consists of two key components: interactive matting with user clicks and uncertainty-guided local refinement. The detailed design will be elaborated in the following sections respectively.

\subsection{Interactive Matting with User Clicks}

Similar to the trimap based matting, besides the original RGB image $I$, we concatenate another one-channel hint heatmap $U$ as the prior input. When the users add a foreground click point, we fill the values within radius $r$ around the click point as $1$ in $U$. Conversely, if a background point is clicked, the corresponding values will be filled with $-1$. All the unspecified values will be filled with 0.

\noindent\textbf{Matting Network.} We design the interactive matting network based on the U-Net architecture\cite{Ronneberger2015UNetCN}, which is widely adopted in existing image matting methods\cite{Tang2019LearningBasedSF,Li2020NaturalIM} and other image processing tasks \cite{he2018deep,zhang2020model}. Specifically, in the encoder part, one stride-$1$ convolution layer and two stride-$2$ convolution layers are first used, then a total of 13 residual blocks consisting of four stages (3,4,4,2) are inserted. In the last three stages, the input resolution will be downsampled by $2$ every time. Symmetrically, the decoder has 10 residual blocks consisting of four stages, followed by one stride-$ \frac{1}{2} $ convolution layer and one stride-$1$ convolution layer. All convolutional layers are followed by batch normalization\cite{Ioffe2015BatchNA} and ReLU except the last output layer. Please refer to the supplementary materials for more network details.

\noindent\textbf{Simulating User Click Interactions.} To make the matting network adapt its behavior to the user clicks, one challenge is collecting such type of training data. This is not only expensive, but also comes with a chicken and egg problem, because the user interaction often depends on the matting output itself. To circumvent this problem, we propose to train the matting network with simulated user interactions. In detail, during training, we randomly sample a total of $m$ click points with radius $r$ on the foreground and background regions for each image, where $m$ is drawn from a geometric distribution with $p=\frac{1}{6}$ and $r=15$. A concern with this simulated user interaction may be the potential domain gap issue. However, from experimental results, we empirically find it works very well. Our observation is that, when providing such user hint points during training, the networks will prefer to leverage them for better convergence and performance but these hints indeed help.

\vspace{0.9em} \noindent\textbf{Loss Functions.} To train the matting network, we adopt two types of loss functions in the image and gradient space respectively:

\begin{equation}
	\begin{aligned}
		\mathcal{L}_{alpha}&=\mathcal{L}_{reg}+\mathcal{L}_{grad}. \\
	\end{aligned}
\end{equation}
For the image space regression loss $\mathcal{L}_{reg}$, it adopts the $ \ell_1 $ loss in semi-transparent regions $T$ and $ \ell_2 $ loss in foreground and background regions $S=\{F,B\}$.
\begin{equation}
	\mathcal{L}_{reg}=\frac{1}{|\mathit{T}|}\sum\limits_{i\in\mathit{T}}|\alpha^{i}_{p}-\alpha^{i}_{g}|+\frac{1}{|\mathit{S}|}\sum\limits_{j\in\mathit{S}}(\alpha^{j}_{p}-\alpha^{j}_{g})^{2},
\end{equation}
where $ \alpha_{p} $ and $ \alpha_{g} $ indicate the predicted and ground-truth alpha values, and $|x|$ denotes the element number of $x$. The gradient loss $ \mathcal{L}_{grad} $ is defined as $ \ell_1 $ loss on the spatial gradient between predicted and ground-truth alpha mattes:
\begin{equation}
	\mathcal{L}_{grad}=\frac{1}{|I|}\sum\limits_{i\in\mathit{\Omega}}|\bigtriangledown(\alpha^{i}_{p})-\bigtriangledown(\alpha^{i}_{g})|,
\end{equation}
where $ \bigtriangledown(x) $ represents the gradient magnitude of $x$. As shown in previous methods\cite{Sengupta2020BackgroundMT,Tang2019LearningBasedSF}, $ \mathcal{L}_{grad} $ can encourage the network to produce sharper matting results.
\subsection{Uncertainty-guided Refinement}
\vspace{0.3em} \noindent\textbf{Uncertainty Estimation Module.} To accomplish uncertainty estimation for the predicted alpha matte $\alpha_p$, we add another similar decoder upon the encoder of the matting network. And we further model alpha matte prediction as a parametric distribution ($p(\alpha|I,U; \mathcal{D})$) learning problem on the whole training set $\mathcal{D}$. Here we adopt the classical univariate Laplace distribution by default:

\begin{equation}
	f(x|\,\mu,\sigma)=\frac{1}{2\sigma}e^{-\frac{|x-\mu|}{\sigma}}, \label{pdfunc}
\end{equation}
In our task, $\mu$ is just the target alpha matte $\alpha_p$ output from the matting network, and $\sigma$ is just the uncertainty prediction $\sigma_p$ output from the uncertainty estimation decoder. To optimize the uncertainty decoder, we use negative log-likelihood minimizing from the probabilistic perspective.

\begin{equation}
	\mathcal{L}_{ue}=-\log p(\alpha|I,U; D)=\frac{1}{|\mathcal{N}|}\sum_{I\in\mathcal{D}}(\mathrm{log}\,\sigma_{p}+\frac{|x-\alpha_{p}|}{\sigma_{p}}),
	\label{un_eq}
\end{equation}
where $\mathcal{N}$ is the total image number of $\mathcal{D}$. During training, the above equation can be directly optimized on the training set by regarding the ground truth alpha matte as $x$. And larger $ \sigma_{p} $ indicates that the network is more uncertain about the output value of matting network.

\noindent\textbf{Local Refinement Network.} After getting the uncertainty map,  we can know where the alpha matting network is not confident and optimize corresponding local parts. In detail, we will crop small $k\times k$ ($k=64$ by default) image patches and the corresponding alpha matte patches for each local part, and then feed them into a small refinement network to get the refined alpha mattes. For the detailed network structure, we simply design a fully convolutional network without downsampling layers. Specifically, besides the two convolutional layers at the beginning and the end, four residual blocks are inserted in the middle part. Since $k$ is often much smaller than the original image size, the computation cost for each local part is smaller. Our method effectively avoids the redundant computation of regions that do not need polishing.

\noindent\textbf{Loss Functions.} Since most pixels in the cropped alpha matte patch are already predicted correctly, only a few "hard" pixels need significant refinement. To make the network pay more attention to those "hard" pixels, we adopt a simplified hard-sample mining objective function $ \mathcal{L}_{refine} $ as follows:
\begin{equation}
	\mathcal{L}_{refine}=\frac{1}{|\mathit{C}|}\sum\limits_{i\in\mathit{C}}|\alpha^{i}_{p}-\alpha^{i}_{g}|+\lambda\frac{1}{|\mathit{H}|}\sum\limits_{j\in\mathit{H}}|\alpha^{j}_{p}-\alpha^{j}_{g}|,
\end{equation}
where $ \mathit{C} $ represents the whole pixel set and $ \mathit{H} $ denotes "hard" pixel set whose error to corresponding ground truth ranks in the top $20\%$ of the entire patch. $\lambda$ denotes the weight that emphasizes the hard samples and set as $1$ by default.

\section{Experiments}
In this section, we will first introduce some implementation details. Then we will evaluate the proposed method on both the synthetic DIM dataset \cite{Xu2017DeepIM} and the real portrait dataset. In addition to the comparison with existing matting methods, detailed ablation analysis is also provided to justify the effects of different components and the special advantages of our method.

\noindent\textbf{Implementation Details.} 
In order to avoid overfitting and improve the generalization ability, we follow GCA-matting \cite{Li2020NaturalIM} and apply different types of data augmentation onto the input images and their corresponding ground truths. Typical augmentation operations include random affine transformation, random cropping, resizing and flipping. Regarding the training strategy, we first train the matting network alone without the uncertainty estimation decoder. After the matting network converges, we freeze it and then train the uncertainty estimation decoder. Empirically, we find this training strategy can significantly boost the training stability. To train the refinement network, we first use the pretrained matting network to predict the alpha matte for the whole training dataset, then compute the absolute prediction error and choose the most challenging patches as training samples. For all the network training, the base learning rate is set to $ 5\times10^{-4} $ with the cosine learning rate scheduler\cite{He2019BagOT}. By default, the matting network is trained for 150 epochs, while the uncertainty estimation decoder and the refinement network are trained for 75 epochs. The Adam optimizer is used, with $ \beta_{1} $ and $ \beta_{2} $ set to $ 0.5 $ and $ 0.999 $ respectively.

\begin{table}[!t]
	\centering

	\begin{tabular}{lcccc}
		\hline
		Methods & SAD & MSE & Grad & Conn \\
		\hline
		CS-Matting & 6.38 & 41.53 & 71.57 & 17.98 \\
		Closed-Form & 7.01 & 45.37 & 80.10 & 24.72 \\
		KNN Matting & 7.79 & 49.94 & 86.26 & 32.78 \\
		Shared Matting & 6.55 & 41.77 & 85.41 & 69.02 \\
		Global Matting & 7.09 & 43.92 & 78.29 & 25.12 \\
		\cdashline{1-5}[1.8pt/2.5pt]
		DIM & 1.93 & 4.18  & 14.20 & 18.92 \\
		CA-Matting & 1.62 & \textbf{3.03}  & 8.90  & 15.16 \\
		GCA-Matting & \textbf{1.50} & 3.19  & 8.97  & \textbf{13.81} \\
		\cdashline{1-5}[1.8pt/2.5pt]
		LF-Matting & 3.47 & 11.69 & 22.50 & 35.74 \\
		\hline
		No-Hints(Ours) & 2.26 & 5.22  & 13.87 & 20.44 \\
		Hints-Train (Ours) & 2.17 & 4.98  & 12.96 & 19.38 \\
		Hints-TrainTest(Ours)  & 1.68 & 3.06  & \textbf{7.57}  & 14.17\\				
		\hline
	\end{tabular}
	\smallskip
	\caption{Quantitative comparison on the DIM dataset. The metrics SAD, MSE, Grad and Conn are scaled by $ 10^{2} $, $ 10^{3} $, $ 10^{5} $ and $ 10^{3} $, respectively. Except LF-Matting, all other methods are trimap-based.}	
	\label{tab:quantity}
\end{table}

\subsection{Evaluation on the DIM dataset}
As the standard matting dataset, DIM contains 43,100 synthetic images for training, which are derived from 431 unique foreground objects and 43,100 background images randomly selected from the MS-COCO dataset\cite{Lin2014MicrosoftCC}. For testing, 50 independent foreground objects and each object is combined with 20 background images from the PASCAL VOC dataset\cite{EveringhamMark2010ThePV} to obtain 1000 synthetic test images. For quantitative evaluation, four popular evaluation metrics proposed in \cite{Rhemann2009APM} are adopted:  the sum of absolute differences (SAD), mean square error (MSE), the gradient (Grad) and connectivity (Conn). For all the metrics, lower indicates better. Empirically, compared to SAD and MSE, Grad and Conn are more representative in terms of the visual quality. 
\begin{figure*}[t]
	\centering
	\includegraphics[width=0.95\textwidth]{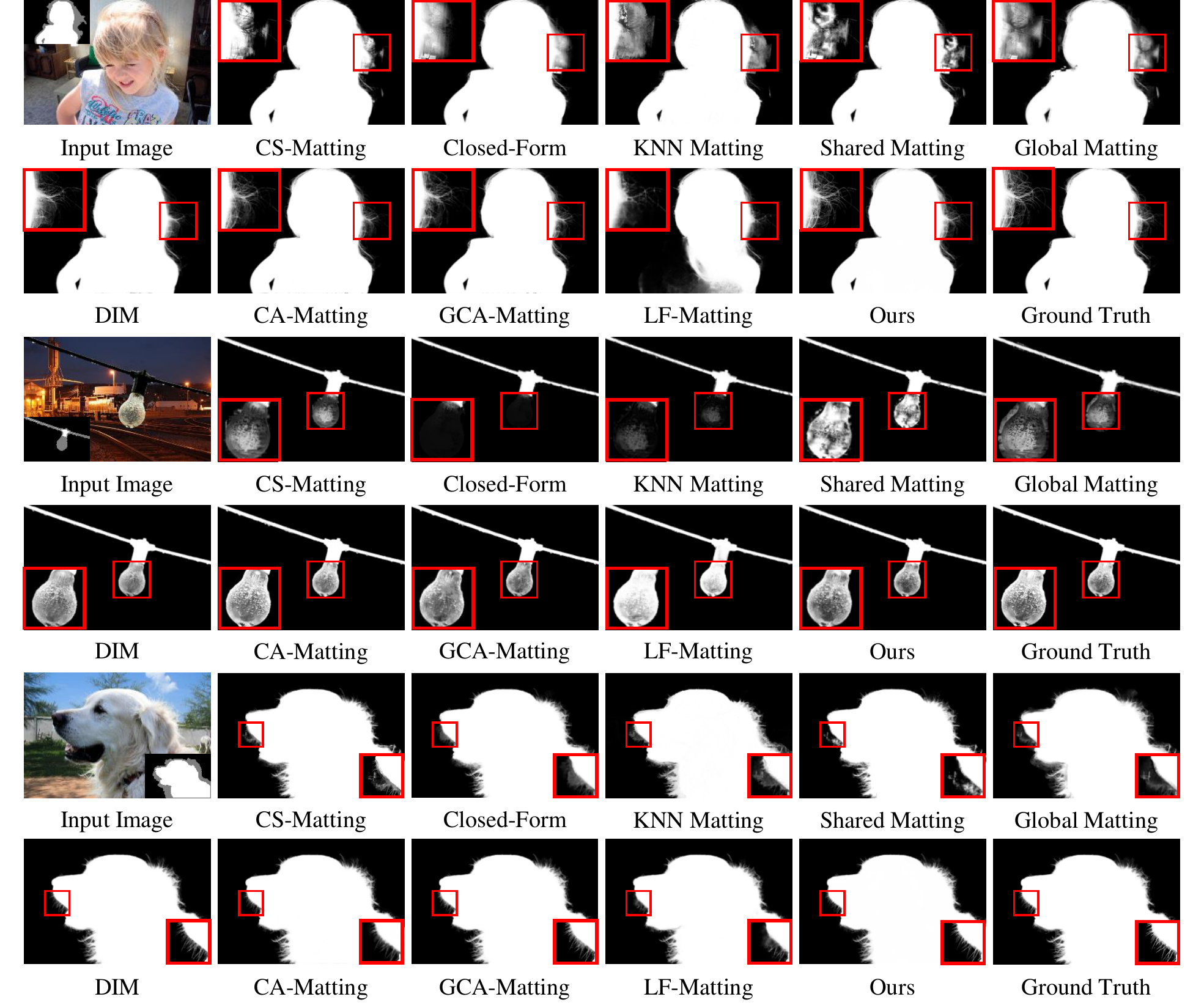} 
	\caption{Visual comparison results on the DIM dataset.}
	\label{fig:qualitative comparison}
\end{figure*}

\noindent\textbf{Quantitative and qualitative comparison.} We compare our approach with many existing matting methods, including traditional methods: CS-Matting\cite{Shahrian2013ImprovingIM}, Closed-Form\cite{Levin2008ACS}, KNN Matting\cite{Chen2013KNNM}, Shared Matting\cite{Gastal2010SharedSF}, Global Matting\cite{He2011AGS}, and learning based methods : DIM\cite{Xu2017DeepIM}, CA-Matting\cite{Hou2019ContextAwareIM}, GCA-Matting\cite{Li2020NaturalIM} and LF-Matting\cite{Zhang2019ALF}. Except LF-Matting, all the remaining methods are trimap-based. And due to the structural constraints of LF-Matting, we keep the aspect ratio of all the test set images while resizing their long sides to 800 pixels. 

\begin{figure}[t]
	\centering
	\includegraphics[width=0.95\columnwidth]{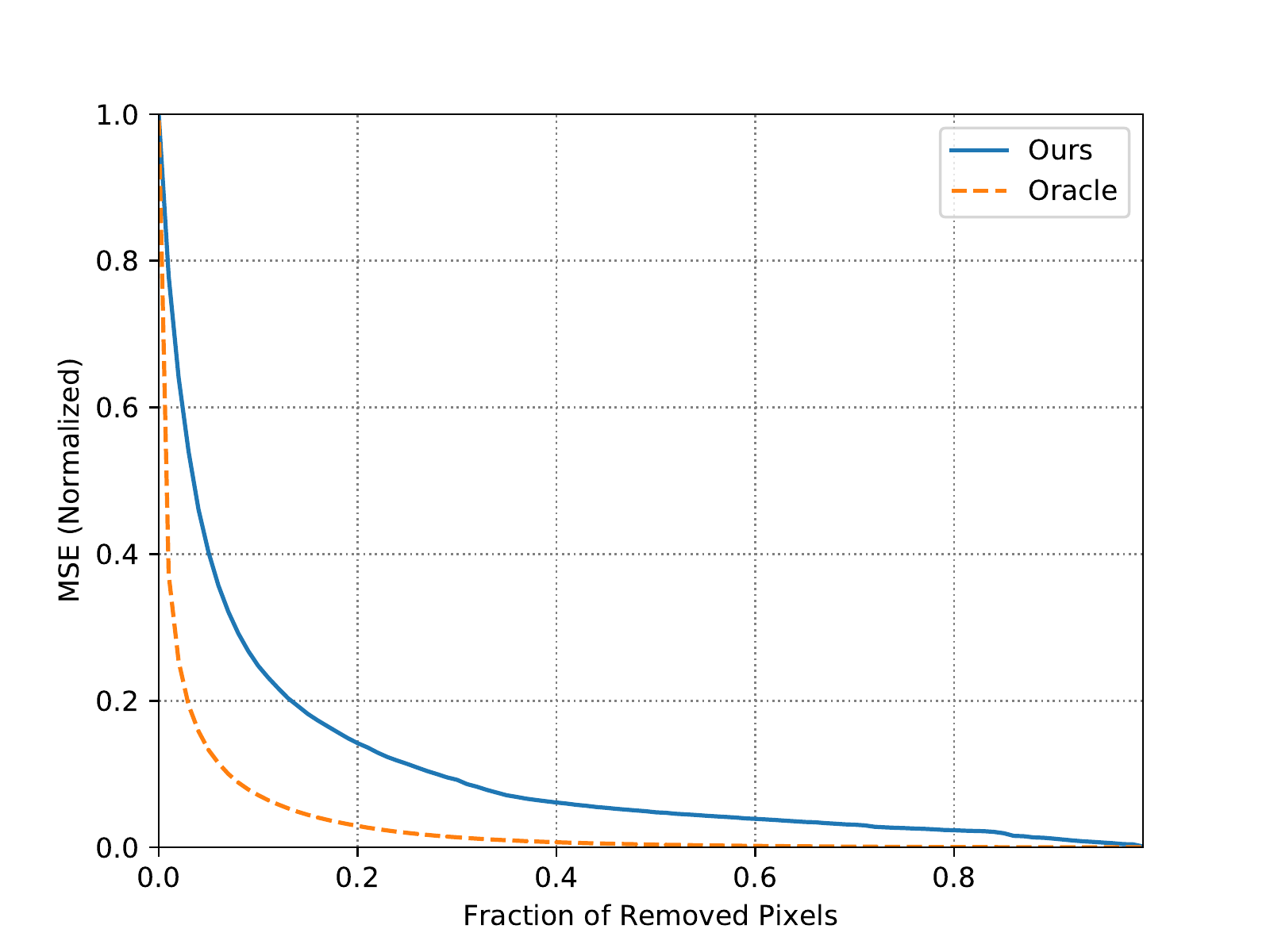} 
	\caption{Sparsification plot of our method for uncertainty  accuracy evaluation. ``Oracle" means the lower boundary by using ground truth error rank.}
	\label{fig:SparsificationPlots}
\end{figure}

As shown in Table \ref{tab:quantity}, our method (``Hints-TrainTest") outperforms all the traditional methods and trimap-free methods significantly. Compared to trimap-based methods, our results are superior to DIM but slightly worse than CA-Matting and GCA-Matting. However, our method is much more user-friendly to end-users. Specifically, on the DIM test dataset, we try to manually draw the trimap for different examples, it will take more than five minutes for each image on average. By contrast, for our method, we only need to add $0.56$ user click point on average, which often just takes several seconds. Besides, the average ``0.56" point also indicates that our method  does not even need any user click for some easy cases. Note that our results shown in Table \ref{tab:quantity} have not used any local refinement. We further provide some qualitative comparison examples in Figure \ref{fig:qualitative comparison}, whose quality rank is consistent with the quantitative results.

 \noindent\textbf{Importance of User Clicks.} To further demonstrate the importance of interactive clicks, we  conduct two baseline experiments \textit{``No-Hints"} and \textit{``Hints-Train"}, and show the results in \Tref{tab:quantity}. In the first baseline, the input is only the original image without any hint prior both in training and testing. And in the second baseline, we only train the network with simulated user click hints but test without any user click. By comparing the results of \textit{``No-Hints"} and \textit{``Hints-Train"}, we can observe that adding user hints into training itself can improve the network performance. It may indicate that the existence of ambiguity will also affect the network training but not just testing. By incorporating a few user clicks during testing, the performance can be further significantly boosted.

 \noindent\textbf{Quality of Uncertainty Estimation.} In order to quantitatively evaluate the uncertainty estimation accuracy, we adopt the widely used Sparsification plots as \cite{Aodha2013LearningAC,Kondermann2008ASC, Ilg2018UncertaintyEA}. Intuitively, Sparsification plots can reveal the degree to which the estimated uncertainty coincides with the true error. As the reference, we also plot the optimal case (``\textit{Oracle}") which represents the lower boundary by gradually removing pixel errors ranked by MSE between predicted alpha matte and the corresponding ground-truth. As shown in Figure \ref{fig:SparsificationPlots},  the predicted uncertainty has a very close curve to \textit{``Oracle"}. By gradually removing the pixel ranked by the predicted uncertainty, the MSE monotonically decreases and its slope becomes smaller. And removing $ 20\% $ of the pixels reduces the MSE by nearly $ 90\% $. Therefore, the estimated uncertainty is a good indicator of the predicted alpha matte quality and guides the users to polish the matting quality based on their computation budget.

\begin{table}[t]
	\centering

	\begin{tabular}{lccccc}
		\hline
		Methods & SAD & MSE & Grad & Conn &Time\\
		\hline
		w/o Ref  & 8.69 & 0.0116 & 5.79 & 8.02 & -\\
		Global Ref  & 8.57 & 0.0117 & 5.81 & 7.89 & 45.00\\
		\hline
		Local Ref-4 & 8.60 & 0.0112 & 5.68 & 7.93 & 2.37\\
		Local Ref-8 & 8.53 & 0.0110 & 5.61 & 7.87 & 4.57\\
		Local Ref-16 & 8.42 & 0.0108 & 5.51 & 7.77 & 10.73\\
		Local Ref-24 & 8.34 & 0.0107 & 5.45 & 7.71 & 20.42\\
		Local Ref-32 & 8.29 & 0.0106 & 5.42 & 7.67 & 31.53\\
		\hline
	\end{tabular}
    \smallskip
	\caption{Quantitative comparison results of refining different number of patches. Ref in the table is an abbreviation for refinement, and we report the time cost in seconds. Obviously, our local refinement mechanism is more effective, efficient and flexible.}	
	\label{tab:ablation}
\end{table}
\begin{figure}[t]
	\centering
	\includegraphics[width=0.95\columnwidth]{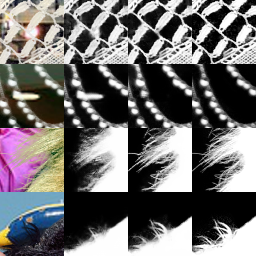} 
	\caption{Visual results before and after uncertainty-guided local refinement. Left to right: original image, results without refinement, results after refinement, ground-truth.}
	\label{fig:localrefine}
\end{figure}

 \noindent\textbf{Importance of Uncertainty-guided Refinement.} To justify the importance of uncertainty-guided local refinement, we selectively refine the top$-K$ local parts guided by the uncertainty map. Since the main goal of the refinement module is polishing the details of transition regions rather than the pure foreground/background regions, we only compute the metrics for transition regions on the DIM dataset for better quantitative evaluation. As a reference, we also give the results of global refinement with the same network structure. As shown in Table \ref{tab:ablation}, by only refining eight $ 64\times64 $ patches, the final alpha matte quality already surpasses the global refinement baseline. As we increase the refinement part number, all four metrics  decrease but with a progressively slower rate, which double demonstrates the accuracy of our estimated uncertainty. Some visual results before and after refinement are shown in Figure \ref{fig:localrefine}. It can be clearly seen that the local refinement part can significantly improve the edge details and remove undesired blurring artifacts. Theoretically, the computation flops ratio between one patch refinement and global refinement is $\frac{k*k}{H*W}$. Therefore, if the local part number to refine is smaller than $\frac{H*W}{k*k}$, our local refinement mechanism is more efficient than the global refinement mechanism. The quantitative time cost comparison results are shown in the last column of \Tref{tab:ablation}. Our local refinements with the different number of patches are faster than the global refinement. Moreover, according to the computation budget, users can customize the number of local patches to be refined, which is not possible with global refinement and shows our local refinement's flexibility.

\subsection{User Study}
To evaluate the convenience of our click interaction, we conducted a user study on the alphamatting.com benchmark test dataset. We compare our approach with a representative method of interaction using scribbles\cite{wang2005iterative}, which also supports trimaps as the auxiliary input. Our user study recruited 10 non-professional participants and provided them with a short training session prior to the user study. Specifically, each user was taught how to use the two interactive systems and how to identify the desired alpha matte of an image.

For click interaction and scribble interaction, each participant was given eight test cases. In each test case, an image from the alphamatting.com test dataset and a reference alpha matte generated by trimap-based GCA-matting\cite{Li2020NaturalIM} were given. The participant was tasked with using both click-based and scribble-based interactive systems to generate an alpha matte that is better or comparable to the reference. The participant can iteratively refine input clicks or scribbles until he or she was satisfied with the alpha matte output. For trimap interaction, the test cases provided to participants also contain trimaps from the alphamatting.com test dataset. We counted the time of the entire interaction process for completing each test case and then calculated the average time among 10 participants and 8 test cases. The average time for the scribble interaction to generate an alpha matte is $ 223 $ seconds and the trimap interaction is $215$ seconds. In contrast, our method takes only $ 46 $ seconds, which is strong evidence that our click-based interaction is more convenient and efficient than the scribble-based or trimap-based ones.

\begin{figure}[t]
	\centering
	\includegraphics[width=0.95\columnwidth]{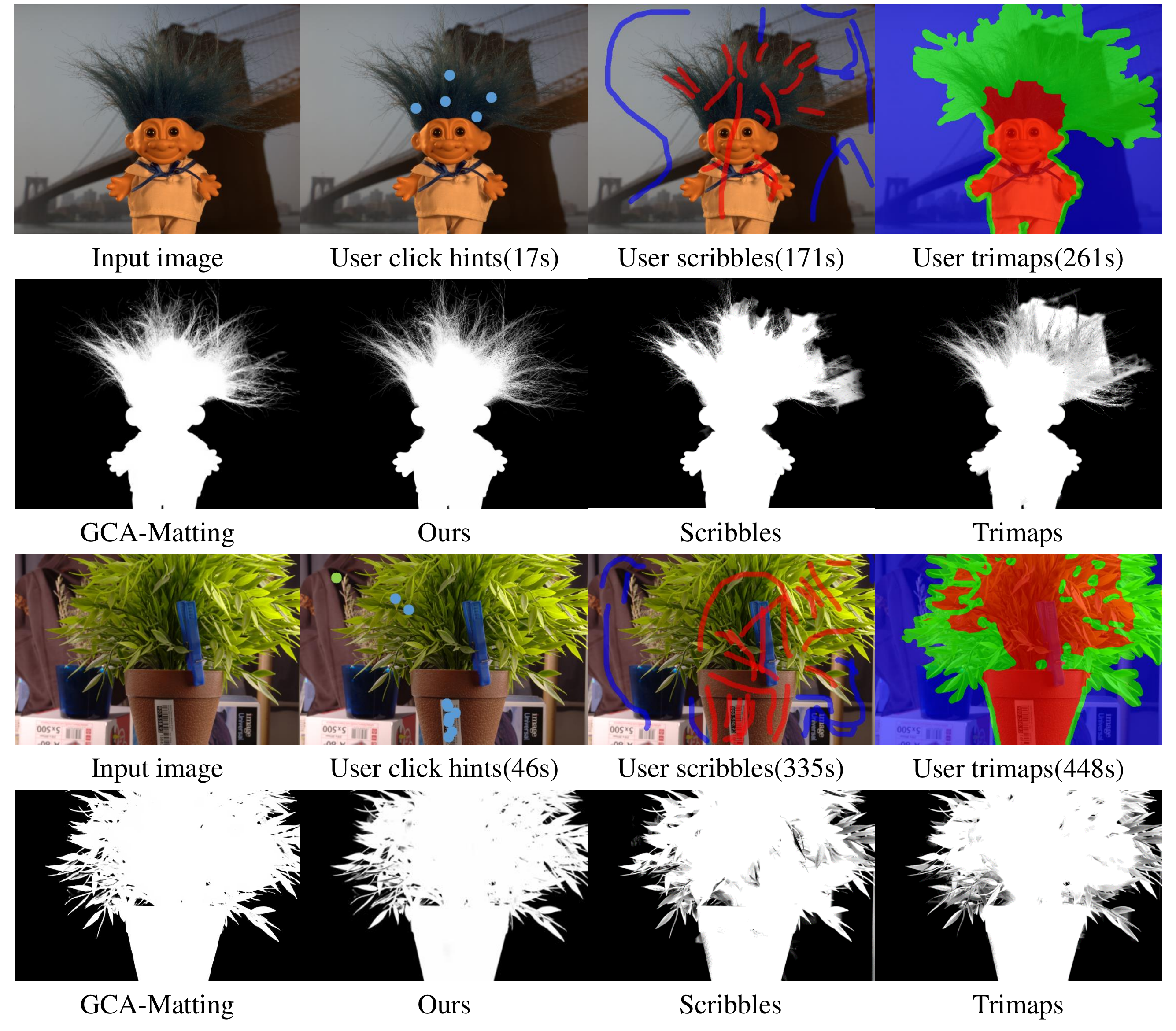} 
	\caption{Selected user study results on the alphamatting.com benchmark test dataset. Since ground truths are not available, we provide state-of-the-art trimap-based GCA-Matting results for comparison. The last three columns show the input images with user inputs, interaction time(in seconds), and results for our method, scribble interaction and trimap interaction, respectively.}
	\label{fig:alphamattingresult}
\end{figure}

We show example results from our user study in \Fref{fig:alphamattingresult}, and the corresponding interaction time is also given. It is clear to perceive that with just a few clicks, our method can achieve comparable matting results to the state-of-the-art trimap-based GCA-matting\cite{Li2020NaturalIM}.

\subsection{Application on the Real Portrait Dataset}
\begin{figure}[t]
	\centering
	\includegraphics[width=0.95\columnwidth]{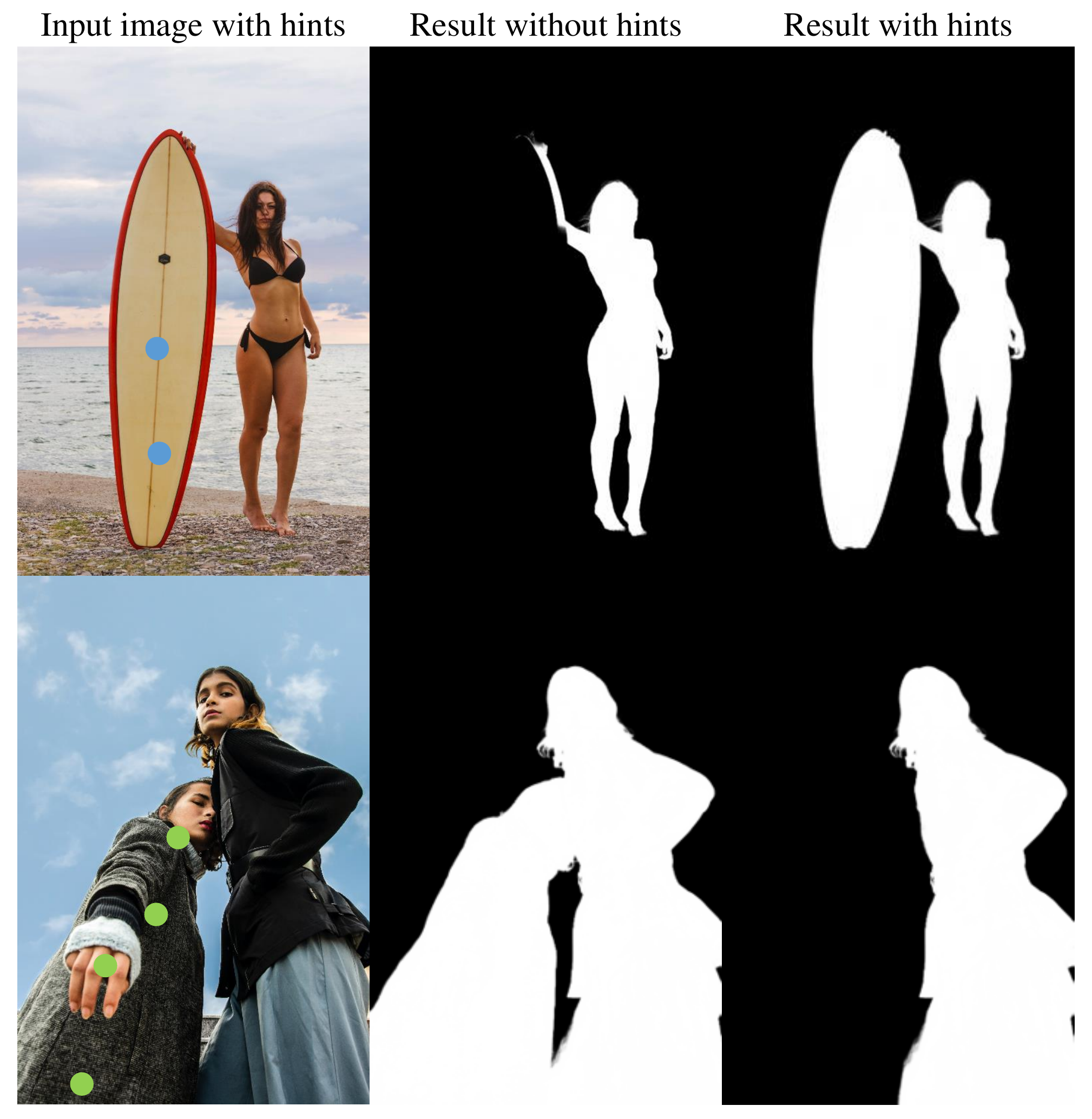} 
	\caption{Visual results on the real dataset. Despite only being trained on the portrait images, our method can easily extend to unseen categories by only providing several user points. By clicking some foreground objects as background, we can also keep only the desired foreground.}
	\label{fig:realimage}
\end{figure}

Besides the synthetic DIM dataset, we further apply our method to real-world portrait images. Specifically, we choose 205 human foreground images from the DIM dataset and about $30k$ images from the human segmentation dataset AISegment\cite{humandataset}. For the DIM dataset, we blend each foreground image with 20 different background images from the MS-COCO dataset, and for the AISegment dataset, we generate pseudo matting labels by using the GCA-Matting network. The trimaps are automatically generated by dilating\&eroding the segmentation masks. And the uncertainty-guided refinement network is only trained with the selected DIM dataset. Since there is no ground truth for the real images, we only provide several visual results here. As shown in \Fref{fig:realimage}, despite only being trained on the portrait category, our method supports matting the objects of new categories by specifying two points on the interested foreground object in the first case. Similarly, it can also remove one specific object with the training category by adding several background points on it. This is a very important feature to remedy existing purely data-driven learning methods.

\section{Conclusion}
In this paper, we propose a new deep image matting framework which supports user click based interaction. Compared to trimap-based methods, our method can achieve comparable performance but with much less user effort. And compared to trimap-free methods, the cheap user click interaction can significantly eliminate the ambiguity and support unseen categories. We also introduce a new uncertainty-guided local refinement mechanism into matting for the first time, which is more flexible and effective than existing global refinement mechanism.

{\small
\bibliographystyle{ieee_fullname}
\bibliography{egbib}
}

\end{document}